\documentclass[11pt,a4paper]{article}
\usepackage{times,latexsym}
\usepackage{url}
\usepackage{CJKutf8}
\usepackage[acceptedWithA]{tacl2018v2}
\usepackage[T2A,OT1]{fontenc}

\usepackage{xspace,mfirstuc,tabulary}

\newif\iftaclinstructions
\taclinstructionsfalse 
\iftaclinstructions
\renewcommand{\confidential}{}
\renewcommand{\anonsubtext}{(No author info supplied here, for consistency with
TACL-submission anonymization requirements)}
\newcommand{\instr}
\fi

\iftaclpubformat 

\else

\fi

\usepackage{amsmath}
\usepackage{cleveref}

\usepackage{amsfonts}
\usepackage{xcolor}

\usepackage[pdftex]{graphicx}

\usepackage{tabu}
\usepackage{booktabs}
\usepackage{multirow}

\usepackage{mathtools}

\DeclarePairedDelimiter{\norm}{\lvert\lvert}{\rvert\rvert}

\usepackage{bm}

\def\eqref#1{equation~\ref{#1}}

\def\1{\bm{1}}

\def\mX{{\bm{X}}}
\def\mY{{\bm{Y}}}

\DeclareMathAlphabet{\mathsfit}{\encodingdefault}{\sfdefault}{m}{sl}
\SetMathAlphabet{\mathsfit}{bold}{\encodingdefault}{\sfdefault}{bx}{n}

\usepackage{algorithm}
\usepackage{algorithmicx}
\usepackage{algpseudocode}

\def\bertsrc{f_{\text{src}}}
\def\berttgt{f_{\text{tgt}}}

\usepackage[normalem]{ulem}

\title{Unsupervised Bitext Mining and Translation \\ via Self-Trained Contextual Embeddings}

\author{Phillip Keung\textsuperscript{•} \quad Julian Salazar\textsuperscript{•} \quad Yichao Lu\textsuperscript{•} \quad Noah A. Smith\textsuperscript{†‡} \\
\textsuperscript{•}Amazon \quad \textsuperscript{†}University of Washington \quad \textsuperscript{‡}Allen Institute for AI \\
{\small \tt \{keung,julsal,yichaolu\}@amazon.com \quad nasmith@cs.washington.edu}
}

\date{}

\begin{document}
\maketitle

\begin{abstract}
We describe an unsupervised method to create \textit{pseudo-parallel corpora} for machine translation (MT) from unaligned text. We use multilingual BERT to create source and target sentence embeddings for nearest-neighbor search and adapt the model via self-training. We validate our technique by extracting parallel sentence pairs on the BUCC 2017 bitext mining task and observe up to a 24.5 point increase (absolute) in $F_1$ scores over previous unsupervised methods. We then improve an XLM-based unsupervised neural MT system pre-trained on Wikipedia by supplementing it with pseudo-parallel text mined from the same corpus, boosting unsupervised translation performance by up to 3.5 BLEU on the WMT'14 French-English and WMT'16 German-English tasks and outperforming the previous state-of-the-art. Finally, we enrich the IWSLT'15 English-Vietnamese corpus with pseudo-parallel Wikipedia sentence pairs, yielding a 1.2 BLEU improvement on the low-resource MT task. We demonstrate that unsupervised bitext mining is an effective way of augmenting MT datasets and complements existing techniques like initializing with pre-trained contextual embeddings.
\end{abstract}

\begin{figure*}
\centering
\includegraphics[width=0.98\linewidth,trim={0 4.0cm 0 5.0cm},clip]{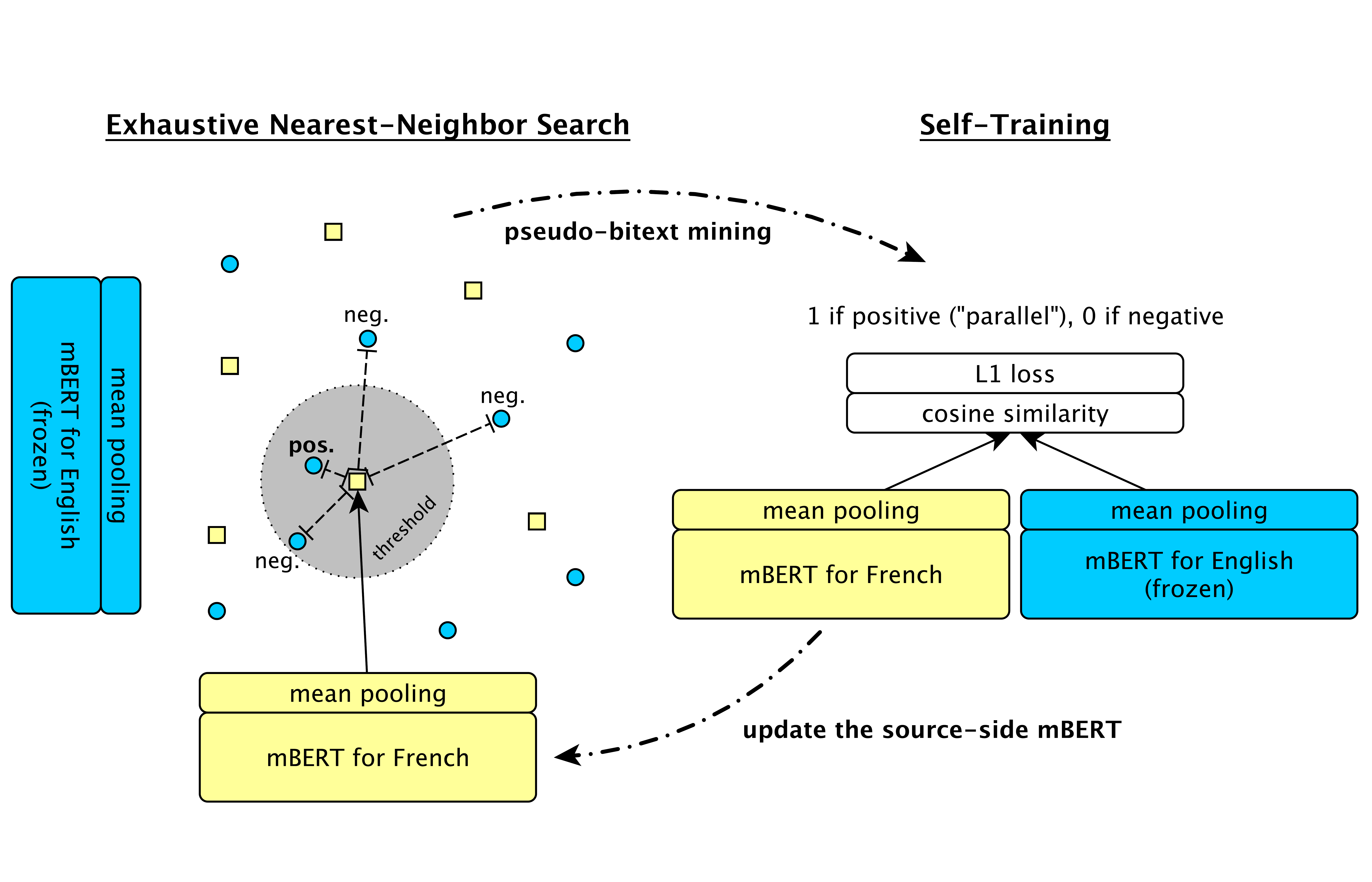}
\caption{Our self-training scheme. \textbf{Left:} We index sentences using our two encoders. For each source sentence, we retrieve $k$ nearest-neighbor target sentences per the margin criterion (Eq.~\ref{margin_func}), depicted here for $k=4$. If the nearest neighbor is within a threshold, it is treated with the source sentence as a positive pair, and the remaining $k-1$ are treated with the source sentence as negative pairs. \textbf{Right:} We refine one of the encoders such that the cosine similarity of the two embeddings is maximized on positive pairs and minimized on negative pairs.}
\label{fig:self-training}
\end{figure*}

\section{Introduction}

Large corpora of parallel sentences are prerequisites for training models across a diverse set of applications, such as neural machine translation (NMT; \citealp{bahdanau-translate}), paraphrase generation \citep{bannard-callison-burch-2005-paraphrasing}, and aligned multilingual sentence embeddings \citep{Artetxe2019LASER}. Systems that extract parallel corpora typically rely on various cross-lingual resources (e.g., bilingual lexicons, parallel corpora), but recent work has shown that unsupervised parallel sentence mining \citep{Hangya2018Unsupervised} and unsupervised NMT \citep{Artetxe2018UNMT,Lample2018UNMT} produce surprisingly good results.\footnote{By \textit{unsupervised}, we mean that no cross-lingual resources like parallel text or bilingual lexicons are used. Unsupervised techniques have been used to bootstrap MT systems for low-resource languages like Khmer and Burmese \citep{marie-etal-2019-supervised}.}

Existing approaches to unsupervised parallel sentence (or \textit{bitext}) mining start from bilingual word embeddings (BWEs) learned via an unsupervised, adversarial approach \citep{Lample2018Word}. \citet{Hangya2018Unsupervised} created sentence representations by mean-pooling BWEs over content words. To disambiguate semantically similar but non-parallel sentences, \citet{hangya-fraser-2019-unsupervised} additionally proposed parallel segment detection by searching for paired substrings with high similarity scores per word. However, using word embeddings to generate sentence embeddings ignores sentential context, which may degrade bitext retrieval performance. 

We describe a new unsupervised bitext mining approach based on contextual embeddings. We create sentence embeddings by mean-pooling the outputs of multilingual BERT (mBERT; \citealp{devlin-etal-2019-bert}), which is pre-trained on unaligned Wikipedia sentences across 104 languages. For a pair of source and target languages, we find candidate translations by using nearest-neighbor search with margin-based similarity scores between pairs of mBERT-embedded source and target sentences. We bootstrap a dataset of positive and negative sentence pairs from these initial neighborhoods of candidates, then self-train mBERT on its own outputs. A final retrieval step gives a corpus of \emph{pseudo-parallel} sentence pairs, which we expect to be a mix of actual translations and semantically-related non-translations.

We apply our technique on the BUCC 2017 parallel sentence mining task \citep{zweigenbaum-etal-2017-overview}. We achieve state-of-the-art $F_1$ scores on unsupervised bitext mining, with an improvement of up to 24.5 points (absolute) on published results \citep{hangya-fraser-2019-unsupervised}. Other work (e.g., \citealp{bert-layer8}) has shown that retrieval performance varies substantially with the layer of mBERT used to generate sentence representations; using the optimal mBERT layer yields an improvement as large as 44.9 points.

Furthermore, our pseudo-parallel text improves unsupervised NMT (UNMT) performance. We build upon the UNMT framework of \citet{lample-etal-2018-phrase} and XLM \citep{Lample2019XLM} by incorporating our pseudo-parallel text (also derived from Wikipedia) at training time. This boosts performance on WMT'14 En-Fr and WMT'16 En-De by up to 3.5 BLEU over the XLM baseline, outperforming the state-of-the-art on unsupervised NMT \citep{Song2019MASS}. 

Finally, we demonstrate the practical value of unsupervised bitext mining in the low-resource setting. We augment the English-Vietnamese corpus (133k pairs) from the IWSLT'15 translation task \cite{cettolo2015iwslt} with our pseudo-bitext from Wikipedia (400k pairs), and observe a 1.2 BLEU increase over the best published model \citep{nguyen2019transformers}. When we reduced the amount of parallel and monolingual Vietnamese data by a factor of ten (13.3k pairs), the model trained with pseudo-bitext performed 7 BLEU points better than a model trained on the reduced parallel text alone.

\section{Our approach}
\label{ssec:approach}
Our aim is to create a bilingual sentence embedding space where, for each source sentence embedding, a sufficiently close nearest neighbor among the target sentence embeddings is its translation. By aligning source and target sentence embeddings in this way, we can extract sentence pairs to create new parallel corpora. \citet{artetxe-schwenk-2019-margin} construct this space by training a joint encoder-decoder MT model over multiple language pairs and using the resulting encoder to generate sentence embeddings. A margin-based similarity score is then computed between embeddings for retrieval (\Cref{ssec:margin}). However, this approach requires large parallel corpora to train the encoder-decoder model in the first place. 

We investigate whether contextualized sentence embeddings created with unaligned text are useful for \emph{unsupervised bitext retrieval}. Previous work explored the use of multilingual sentence encoders taken from machine translation models (e.g., \citealp{Artetxe2019LASER, lu2018neural}) for zero-shot cross-lingual transfer. Our work is motivated by recent success in tasks like zero-shot text classification and named entity recognition (e.g., \citealp{keung-etal-2019-adversarial, mulcaire2019polyglot}) with multilingual contextual embeddings, which exhibit cross-lingual properties despite being trained without parallel sentences.

We illustrate our method in \Cref{fig:self-training}. We first retrieve the candidate translation pairs:
\begin{itemize}
\itemsep0em 
    \item Each source and target language sentence is converted into an embedding vector with mBERT via mean-pooling.
    \item Margin-based scores are computed for each sentence pair using the $k$ nearest neighbors of the source and target sentences (Sec.~\ref{ssec:margin}).
    \item Each source sentence is paired with its nearest neighbor in the target language based on this score.
    \item We select a threshold score that keeps some top percentage of pairs (Sec.~\ref{ssec:margin}).
    \item Rule-based filters are applied to further remove mismatched sentence pairs (Sec.~\ref{ssec:filtering}).
\end{itemize}

The remaining candidate pairs are used to bootstrap a dataset for self-training mBERT as follows:
\begin{itemize}
\itemsep0em 
    \item Each candidate pair (a source sentence and its closest nearest neighbor above the threshold) is taken as a positive example.
    \item This source sentence is also paired with its next $k-1$ neighbors to give hard negative examples (we compare this with random negative samples in Sec.~\ref{ssec:negative}).
    \item We finetune mBERT to produce sentence embeddings that discriminate between positive and negative pairs (Sec.~\ref{ssec:self-training}).
\end{itemize}

After self-training, the finetuned mBERT model is used to generate new sentence embeddings. Parallel sentences should be closer to each other in this new embedding space, which improves retrieval performance.

\subsection{Sentence embeddings and nearest-neighbor search}

We use mBERT \citep{devlin-etal-2019-bert} to create sentence embeddings for both languages by mean-pooling the representations from the final layer. We use FAISS \citep{Johnson2017FAISS} to perform exact nearest-neighbor search on the embeddings. We compare every sentence in the source language to every sentence in the target language; we do not use links between Wikipedia articles or other metadata to reduce the size of the search space. In our experiments, we retrieve the $k = 4$ closest target sentences for each source sentence; the source language is always non-English, while the target language is always English.

\subsection{Margin-based score}
\label{ssec:margin}

We compute a margin-based similarity score between each source sentence and its $k$ nearest target neighbors. Following \citet{artetxe-schwenk-2019-margin}, we use the \emph{ratio} margin score, which calibrates the cosine similarity by dividing it by the average cosine distance of each embedding's $k$ nearest neighbors:
\begin{align}
\begin{split}
	&\textrm{margin}(x,y) = \\
	&\frac{\cos(x,y)}{\sum_{z \in \text{NN}^{\text{tgt}}_k(x)} \frac{\cos(x,z)}{2k} + \sum_{z \in \text{NN}^{\text{src}}_k(y)} \frac{\cos(y,z)}{2k}}.
\end{split}
\label{margin_func}
\end{align}
We remove the sentence pairs with margin scores below some pre-selected threshold. For BUCC, we do not have development data for tuning the threshold hyperparameter, so we simply use the prior probability. For example, the creators of the dataset estimate that $\sim$2\% of De sentences have an En translation, so we choose a score threshold such that we retrieve $\sim$2\% of the pairs. We set the threshold in the same way for the other BUCC pairs. For UNMT with Wikipedia bitext mining, we set the threshold such that we always retrieve 2.5 million sentence pairs for each language pair.

\subsection{Rule-based filtering}
\label{ssec:filtering}

\begin{table*}[h!]
\begin{minipage}{1.0\linewidth}
	\centering
	\footnotesize
\begin{tabu}{lcccc}
\toprule
\textbf{Method} & \textbf{De-En} & \textbf{Fr-En} & \textbf{Ru-En} & \textbf{Zh-En} \\
  \midrule 
  \scriptsize \textit{Hangya and Fraser (2019)} \\
  \midrule 
 avg.\ & 30.96 & 44.81 & 19.80 & - \\
 align-static & 42.81 & 42.21 & 24.53 & - \\
 align-dyn.\ & 43.35 & 43.44 & 24.97 & - \\
\midrule
\scriptsize \textit{Our method} \\
\midrule 
mBERT (final layer) & 42.1 & 45.8 & 36.9 & 35.8 \\
+ digit filtering (DF) & 47.0 & 49.3 & 41.2 & 38.0 \\
+ edit distance (ED) & 47.0 & 49.3 & 41.2 & 38.0 \\
+ self-training (ST) & \textbf{60.6} & \textbf{60.2} & \textbf{49.5} & \textbf{45.7} \\
\midrule 
mBERT (layer 8) & 67.0 & 65.3 & 59.3 & 53.3 \\
+ DF, ED, ST & \textbf{74.9} & \textbf{73.0} & \textbf{69.9} & \textbf{60.1} \\
\bottomrule
\end{tabu}
\end{minipage}
\caption{$F_1$ scores for unsupervised bitext retrieval on BUCC 2017. Results with mBERT are from our method (Sec.~\ref{ssec:approach}) using the final (12th) layer. We also include results for the 8th layer (e.g., \citealp{bert-layer8}), but do not consider this part of the unsupervised setting as we would not have known \emph{a priori} which layer was best to use.}
\label{table:bucc-2017}
\end{table*}

We also apply two simple filtering steps before finalizing the candidate pairs list:
\begin{itemize}
	\item \textbf{Digit filtering}: Sentence pairs which are translations of each other must have digit sequences that match exactly.\footnote{In Python, \tt{set(re.findall("[0-9]+",sent1)) == set(re.findall("[0-9]+",sent2))}.}
	\item \textbf{Edit distance}: Sentences from English Wikipedia sometimes appear in non-English pages and vice versa. We remove sentence pairs where the content of the source and target share substantial overlap (i.e., the character-level edit distance is $\leq$50\%).
\end{itemize}

\subsection{Self-training}
\label{ssec:self-training}

We devise an unsupervised self-training technique to improve mBERT for bitext retrieval using mBERT's own outputs. For each source sentence, if the nearest target sentence is within the threshold and not filtered out, the pair is treated as a positive sentence. We then keep the next $k-1$ nearest neighbors as negative sentences. Altogether, these give us a training set of examples which are labeled as positive or negative pairs.

We train mBERT to discriminate between positive and negative sentence pairs as a binary classification task. We distinguish the mBERT encoders for the source and target languages as $\bertsrc$, $\berttgt$ respectively. Our training objective is
\begin{align}
\begin{split}
&\mathcal{L}(\mX, \mY; \Theta_{\text{src}}) =\\
&\left|\frac{\bertsrc(\mX; \Theta_{\text{src}})^\top \berttgt(\mY)}{\norm{\bertsrc(\mX; \Theta_{\text{src}})}\norm{\berttgt(\mY)}} - \textrm{Par}(\mX,\mY)\right|,
\end{split}
\end{align}
where $\bertsrc(\mX)$ and $\berttgt(\mY)$ are the mean-pooled representations of the source sentence $\mX$ and target sentence $\mY$, and where $\textrm{Par}(\mX,\mY)$ is 1 if $\mX,\mY$ are parallel and 0 otherwise. This loss encourages the cosine similarity between the source and target embeddings to increase for positive pairs and decrease otherwise. The process is depicted in \Cref{fig:self-training}.

Note that we only finetune $\bertsrc$ (parameters $\Theta_{\text{src}}$) and we hold $\berttgt$ fixed. If both $\bertsrc$ and $\berttgt$ are updated, then the training process collapses to a trivial solution, since the model will map all pseudo-parallel pairs to one representation and all non-parallel pairs to another. We hold $\berttgt$ fixed, which forces $\bertsrc$ to align its outputs to the target (in our experiments, always English) mBERT embeddings.

After finetuning, we use the updated $\bertsrc$ to generate new non-English sentence embeddings. We then repeat the retrieval process with FAISS, yielding a final set of pseudo-parallel pairs after thresholding and filtering.

\section{Unsupervised bitext mining}
\label{sec:experiments}

We apply our method to the BUCC 2017 shared task, ``Spotting Parallel Sentences in Comparable Corpora'' \citep{zweigenbaum-etal-2017-overview}. The task involves retrieving parallel sentences from monolingual corpora derived from Wikipedia. Parallel sentences were inserted into the corpora in a contextually-appropriate manner by the task organizers. The shared task assessed retrieval systems for precision, recall, and $F_1$-score on four language pairs: De-En, Fr-En, Ru-En, and Zh-En. Prior work on unsupervised bitext mining  has generally studied the European language pairs to avoid dealing with Chinese word segmentation \citep{Hangya2018Unsupervised, hangya-fraser-2019-unsupervised}.

\begin{table*}
    \centering
    \footnotesize
    \begin{tabular}{lp{375pt}}
        \toprule
        \textbf{Language pair} & \textbf{Parallel sentence pair} \\
        
        \midrule
        
        \multirow{2}{*}{De-En} & Beide Elemente des amerikanischen Traums haben heute einen Teil ihrer Anziehungskraft verloren. \\
        & Both elements of the American dream have now lost something of their appeal. \\
        \midrule
        \multirow{2}{*}{Fr-En} & L'Allemagne à elle seule s’attend à recevoir pas moins d'un million de demandeurs d'asile cette année. \\
        & Germany alone expects as many as a million asylum-seekers this year. \\
        \midrule
        \multirow{4}{*}{Ru-En} & {\fontencoding{T2A}\selectfont Однако по решению Берлинского конгресса в 1881 году к территории Греции присоединилась Фессалия и часть Эпира.} \\
        & Nevertheless, in 1881, Thessaly and small parts of Epirus were ceded to Greece as part of the Treaty of Berlin. \\
        \midrule
        \multirow{2}{*}{Zh-En} & \begin{CJK*}{UTF8}{gbsn}在如今这个奇怪的新世界里，现代和前现代相互依存。\end{CJK*} \\
        & In the strange new world of today, the modern and the pre-modern depend on each other. \\
        \bottomrule
    \end{tabular}
        \caption{Examples of parallel sentences that were extracted by our  method on the BUCC 2017 shared task.}
    \label{table:examples}
\end{table*}

\subsection{Setup}

For each BUCC language pair, we take the corresponding source and target monolingual corpus, which have been pre-split into \emph{training}, \emph{sample}, and \emph{test} sets at a ratio of 49\%--2\%--49\%. The identity of the parallel sentence pairs for the test set were not publicly released, and are only available for the training set. Following the convention established in \citet{hangya-fraser-2019-unsupervised} and \citet{artetxe-schwenk-2019-margin}, we use the \emph{test} portion for unsupervised system development and evaluate on the \emph{training} portion.

We use the reference FAISS implementation\footnote{\url{https://github.com/facebookresearch/faiss}} for nearest-neighbor search. We used the GluonNLP toolkit \citep{Guo2019GluonNLP} with pre-trained mBERT weights\footnote{\url{https://github.com/google-research/bert/blob/master/multilingual.md}} for inference and self-training. We compute the margin similarity score in Eq. \ref{margin_func} with $k=4$ nearest neighbors. We set a threshold on the score such that we retrieve the prior proportion (e.g., $\sim$2\%) of parallel pairs in each language. 

We then finetune mBERT via self-training. We take minibatches of 100 sentence pairs. We use the Adam optimizer with a constant learning rate of 0.00001 for 2 epochs. To avoid noisy translations, we finetune on the top 50\% of the highest-scoring pairs from the retrieved bitext (e.g., if the prior proportion is 2\%, then we would use the top 1\% of sentence pairs for self-training).

We considered performing more than one round of self-training but found it was not helpful for the BUCC task. BUCC has very few parallel pairs (e.g., 9,000 pairs for Fr-En) per language and thus few positive pairs for our unsupervised method to find. The size of the self-training corpus is limited by the proportion of parallel sentences, and mBERT rapidly overfits to small datasets.

\subsection{Results}

We provide a few examples of the bitext we retrieved in \Cref{table:examples}. The examples were chosen from the high-scoring pairs and verified to be correct translations.

Our retrieval results are in \Cref{table:bucc-2017}. We compare our results with strictly unsupervised techniques, which do not use bilingual lexicons, parallel text, or other cross-lingual resources. Using mBERT as-is with the margin-based score works reasonably well, giving $F_1$ scores in the range of 35.8 to 45.8, which is competitive with the previous state-of-the-art for some pairs, and outperforming by 12 points in the case of Ru-En. Furthermore, applying simple rule-based filters (Sec.~\ref{ssec:filtering}) on the candidate translation pairs adds a few more points, although the edit distance filter has a negligible effect when compared with the digit filter.

We see that finetuning mBERT on its own chosen sentence pairs (i.e., unsupervised self-training) yields significant improvements, adding another 8 to 14 points to the $F_1$ score on top of filtering. In all, these $F_1$ scores represent a 34\% to 98\% relative improvement over existing techniques in unsupervised parallel sentence extraction for these language pairs.

\citet{bert-layer8} explored bitext mining with mBERT in the supervised context and found that retrieval performance significantly varies with the mBERT layer used to create sentence embeddings. In particular, they found layer 8 embeddings gave the highest precision-at-1. We also observe an improvement (\Cref{table:bucc-2017}) in unsupervised retrieval of another 13 to 20 points by using the 8th layer instead of the default final layer (12th). We include these results but do not consider them unsupervised, as we would not know \textit{a priori} which layer was best to use.

\subsection{Choosing negative sentence pairs}
\label{ssec:negative}

Other authors (e.g., \citealp{guo-etal-2018-effective}) have noted that the choice of negative examples has a considerable impact on metric learning. Specifically, using negative examples which are difficult to distinguish from the positive nearest neighbor is often beneficial for performance. We examine the impact of taking random sentences instead of the remaining $k-1$ nearest neighbors as the negatives during self-training.

Our results are in \Cref{table:negative-type}. While self-training with random negatives still greatly improves the untuned baseline, the use of hard negative examples mined from the $k$-nearest neighborhood can make a significant difference to the final $F_1$ score.

\begin{table}[h!]
\begin{minipage}{1.0\linewidth}
	\centering
	\footnotesize
\begin{tabu}{lcccc}
\toprule
\textbf{Method} & \textbf{De-En} & \textbf{Fr-En} & \textbf{Ru-En} & \textbf{Zh-En} \\
\midrule
mBERT w/o ST & 47.0 & 49.3 & 41.2 & 38.0 \\ 
\midrule
w/ ST (random) & 57.7 & 55.7 & 48.1 & 45.2 \\
w/ ST (hard) & \textbf{60.6} & \textbf{60.2} & \textbf{49.5} & \textbf{45.7} \\
\bottomrule
\end{tabu}
\end{minipage}
\caption{$F_1$ scores for bitext retrieval on BUCC 2017 using random sentences as negative samples instead of nearest neighbors.}
\label{table:negative-type}
\end{table}

\section{Bitext for neural machine translation}
\label{sec:unmt}

A major application of bitext mining is to create new corpora for machine translation. We conduct an extrinsic evaluation of our unsupervised bitext mining approach on \emph{unsupervised} (WMT'14 French-English, WMT'16 German-English) and \emph{low-resource} (IWSLT'15 English-Vietnamese) translation tasks.

We perform large-scale unsupervised bitext extraction on the October 2019 Wikipedia dumps in various languages. We use \texttt{wikifil.pl}\footnote{\url{https://github.com/facebookresearch/fastText/blob/master/wikifil.pl}} to extract paragraphs from Wikipedia and remove markup. We then use the \texttt{syntok}\footnote{\url{https://github.com/fnl/syntok}} package for sentence segmentation. Finally, we reduce the size of the corpus by removing sentences that aren't part of the body of Wikipedia pages. Sentences that contain \texttt{*}, \texttt{=}, \texttt{//}, \texttt{::}, \texttt{\#}, \texttt{www}, \texttt{(talk)}, or the pattern \texttt{[0-9]\{2\}:[0-9]\{2\}} are filtered out. 

We index, retrieve, and filter candidate sentence pairs with the procedure in Sec.~\ref{sec:experiments}. Unlike BUCC, the Wikipedia dataset does not fit in GPU memory. The processed corpus is quite large, with 133 million, 67 million, 36 million, and 6 million sentences in English, German, French, and Vietnamese respectively. We therefore shard the dataset into chunks of 32,768 sentences and perform nearest-neighbor comparisons in chunks for each language pair. We use a simple map-reduce algorithm to merge the intermediate results back together.

We follow the approach outlined in Sec.~\ref{ssec:approach} for Wikipedia bitext mining. For each source sentence, we retrieve the 4 nearest target neighbors across the millions of sentences that we extracted from Wikipedia and compute the margin-based scores for each pair. 

\begin{table*}[h!]
\begin{minipage}{1.0\linewidth}
	\centering
	\footnotesize
\begin{tabu}{llllcccc}
\toprule
\textbf{Reference} & \textbf{Architecture} & \textbf{Pre-training} &  \textbf{En-De} & \textbf{De-En} & \textbf{En-Fr} & \textbf{Fr-En} \\
\midrule
\citet{Artetxe2018UNMT} & 2-layer RNN & & 6.89 & 10.16 & 15.13 & 15.56 \\
\citet{Lample2018UNMT} & 3-layer RNN & & 9.75 & 13.33 & 15.05 & 14.31 \\
\citet{yang-etal-2018-unsupervised} & 4-layer Transformer & & 10.86 & 14.62 & 16.97 & 15.58 \\
\citet{lample-etal-2018-phrase} & 4-layer Transformer & & 17.16 & 21.00 & 25.14 & 24.18 \\
\citet{Song2019MASS} & 6-layer Transformer & MASS & 28.3 & 35.2 & 37.5 & 34.9 \\
\midrule
\scriptsize \textit{XLM Baselines} \\
\midrule 
\citet{Lample2019XLM} & 6-layer Transformer & XLM & -- & -- & 33.4 & 33.3 \\
\citet{Song2019MASS} & 6-layer Transformer & XLM & 27.0 & 34.3 & 33.4 & 33.3 \\
XLM reference implementation & 6-layer Transformer & XLM & -- & -- & 36.6 & 34.0 \\
\emph{Maximum performance across baselines} & 6-layer Transformer & XLM & 27.0 & 34.3 & 36.6 & 34.0 \\
\midrule
\scriptsize \textit{Ours} \\
\midrule 
Our XLM baseline & 6-layer Transformer & XLM & 27.7 & 34.5 & 36.7 & 34.5 \\
w/ pseudo-parallel text before ST & 6-layer Transformer & XLM & 30.4 & 36.3 & 39.7 & 35.9 \\
w/ pseudo-parallel text after ST & 6-layer Transformer & XLM & \textbf{30.7} & \textbf{37.3} & \textbf{40.2} & \textbf{36.9} \\
\bottomrule
\end{tabu}
\end{minipage}
\caption{BLEU scores for unsupervised NMT performance on WMT'14 English-French and WMT'16 English-German test sets. All methods only use unaligned Wikipedia corpora for pre-training and/or bitext mining. `ST' refers to self-training.}
\label{table:unmt}
\end{table*}

\subsection{Unsupervised NMT}

We show that our pseudo-parallel text can complement existing techniques for unsupervised translation \citep{Artetxe2018UNMT, lample-etal-2018-phrase}. In line with existing work on UNMT, we evaluate our approach on the WMT'14 Fr-En and WMT'16 De-En test sets.

Our UNMT experiments build upon the reference implementation\footnote{\label{xlm}\url{https://github.com/facebookresearch/xlm}} of XLM \citep{Lample2019XLM}. The UNMT model is trained by alternating between two steps: a denoising auto-encoder step and a backtranslation step (refer to \citet{lample-etal-2018-phrase} for more details). The backtranslation step generates pseudo-parallel training data, and we incorporate our bitext during UNMT training in the same way, as another set of pseudo-parallel sentences. We also use the same initialization as \citet{Lample2019XLM}, where the UNMT models have encoders and decoders that are initialized with contextual embeddings trained on the source and target language Wikipedia corpora with the masked language model (MLM) objective; no parallel data is used.

We performed the exhaustive (Fr Wiki)-(En Wiki) and (De Wiki)-(En Wiki) nearest-neighbor comparison on eight V100 GPUs, which requires 3 to 4 days to complete per language pair. We retained the top 2.5 million pseudo-parallel Fr-En and De-En sentence pairs after mining.

\subsection{Results}
\label{ssec:unmt-results}

Our results are in \Cref{table:unmt}. The addition of mined bitext consistently increases the BLEU score in both directions for WMT'14 Fr-En and WMT'16 De-En. Much of the existing work on improving UNMT focuses on improved initialization with contextual embeddings like XLM or MASS \citep{Song2019MASS}. These embeddings were already pre-trained on Wikipedia data, so it is surprising that adding our pseudo-parallel Wikipedia sentences leads to a 2 to 3 BLEU improvement. In other words, our approach is complementary to pre-trained initialization techniques.

Previously (in Table \ref{table:bucc-2017}), we saw that self-training improved the $F_1$ score for BUCC bitext retrieval. The improvement in bitext quality carries over to UNMT, and providing better pseudo-parallel text yields a consistent improvement for all translation directions.

Our results are state-of-the-art in UNMT, but they should be interpreted relative to the strength of our XLM baseline. We are building on top of the XLM initialization, and the effectiveness of the initialization (and the various hyperparameters used during training and decoding) affects the strength of our final results. For example, we adjusted the beam width on our XLM baselines to attain BLEU scores which are similar to what others have published. One can apply our method to MASS, which performs better than XLM on UNMT, but we chose to report results on XLM because it has been validated on a wider range of tasks and languages.

We also trained a standard 6-layer transformer encoder-decoder model directly on the pseudo-parallel text. We used the standard implementation in Sockeye \citep{hieber2018sockeye} as-is, and trained models for French and German on 2.5 million Wikipedia sentence pairs. We withheld 10k pseudo-parallel pairs per language pair to serve as a development set. We achieved BLEU scores of 20.8, 21.1, 28.2, and 28.0 on En-De, De-En, En-Fr, and Fr-En respectively. BLEU scores were computed with SacreBLEU \citep{post-2018-call}. This compares favorably with the best UNMT results in \citet{lample-etal-2018-phrase}, while avoiding the use of parallel development data altogether.

\subsection{Low-resource NMT}
\label{ssec:low-resource}

French and German are high-resource languages and are linguistically close to English. We therefore evaluate our mined bitext on a low-resource, linguistically distant language pair. The IWSLT'15 English-Vietnamese MT task \citep{cettolo2015iwslt} provides 133k sentence pairs derived from translated TED talks transcripts and is a common benchmark for low-resource MT. We take supervised training data from the IWSLT task and augment it with different amounts of pseudo-parallel text mined from English and Vietnamese Wikipedia. Furthermore, we construct a very low-resource setting by downsampling the parallel text and monolingual Vietnamese Wikipedia text by a factor of ten (13.3k sentence pairs).

We use the reference implementation\footnote{\url{https://github.com/tnq177/transformers_without_tears}} for the state-of-the-art model \citep{nguyen2019transformers}, which is a highly regularized 6+6-layer transformer with pre-norm residual connections, scale normalization, and normalized word embeddings. We use the same hyperparameters (except for the dropout rate) but train on our augmented datasets. To mitigate domain shift, we finetune the best checkpoint for 75k more steps using only the IWSLT training data, in the spirit of ``trivial'' transfer learning for low-resource NMT \citep{kocmi-bojar-2018-trivial}.

In Table \ref{table:low-resource}, we show BLEU scores as more pseudo-parallel text is included during training. As in previous works on En-Vi (cf.\ \citealp{Luong2015Stanford}), we use tst2012 (1553 pairs) and tst2013 (1268 pairs) as our development and test sets respectively, we tokenize all data with Moses, and we report tokenized BLEU via \texttt{multi-bleu.perl}. The BLEU score increases monotonically with the size of the pseudo-parallel corpus and exceeds the state-of-the-art system's BLEU by 1.2 points. This result is consistent with improvements observed with other types of monolingual data augmentation like pre-trained UNMT initialization, various forms of back-translation \citep{hoang2018iterative,Zhou2020ImprovingNN}, and cross-view training (CVT; \citealp{clark2018semi}):

\begin{table}[h!]
	\centering
	\footnotesize
\begin{tabu}{lc}
\toprule
 & \textbf{En-Vi} \\
\midrule
\citet{Luong2015Stanford} & 26.4 \\
\citet{clark2018semi} & 28.9 \\
\citet{clark2018semi}, with CVT & 29.6 \\
\citet{Xu2019Understanding} & 31.4 \\
\citet{nguyen2019transformers} & 32.8 (28.8) \\
\midrule
+ top 100k mined pairs & 33.2 (29.5) \\
+ top 200k mined pairs & 33.9 (29.8) \\
+ top 300k mined pairs & \textbf{34.0 (30.0)} \\
+ top 400k mined pairs & 34.1 (29.9) \\
\bottomrule
\end{tabu}
\caption{Tokenized BLEU scores on tst2013 for the low-resource IWSLT'15 English-Vietnamese translation task using bitext mined with our method. Added pairs are sorted by their score. Development scores on tst2012 in parentheses.}
\label{table:low-resource}
\end{table}

We describe our hyperparameter tuning and infrastructure following \citet{dodge-etal-2019-show}. The translation sections of this work mostly used default parameters, but we did tune the dropout rate (at 0.2 and 0.3) for each amount of mined bitext for the supervised En-Vi task (at 100k, 200k, 300k and 400k sentence pairs). We include development scores for our best models; dropout of 0.3 did best for 0k and 100k, while 0.2 did best otherwise. Training takes less than a day on one V100 GPU.

To simulate a very low-resource task, we use one-tenth of the training data by downsampling the IWSLT En-Vi train set to 13.3k sentence pairs. Furthermore, we mine bitext from one-tenth of the monolingual Wiki Vi text and extract proportionately fewer sentence pairs (i.e., 10k, 20k, 30k and 40k pairs). We use the implementation and hyperparameters for the regularized 4+4-layer transformer used by \citet{nguyen2019transformers} in a similar setting. We tune the dropout rate (0.2, 0.3, 0.4) to maximize development performance; 0.4 was best for 0k, 0.3 for 10k and 20k, and 0.2 for 30k and 40k. In Table \ref{table:very-low-resource}, we see larger improvements in BLEU (4+ points) for the same relative increases in mined data (as compared to Table \ref{table:low-resource}). In both cases, the rate of improvement tapers off as the quality and relative quantity of mined pairs degrades at each increase.

\begin{table}[h!]
	\centering
	\footnotesize
\begin{tabu}{lc}
\toprule
 & \textbf{En-Vi, one-tenth} \\
\midrule
13.3k pairs (from 133k original) & 20.7 (19.5) \\
+ top 10k mined pairs & 25.0 (22.9) \\
+ top 20k mined pairs & 26.7 (24.1) \\
+ top 30k mined pairs & 27.3 (24.5) \\
+ top 40k mined pairs & \textbf{27.7 (24.7)} \\
\bottomrule
\end{tabu}
\caption{Tokenized BLEU scores (tst2013), where the bitext was mined from one-tenth of the monolingual Vietnamese data. Development scores on tst2012 in parentheses.}
\label{table:very-low-resource}
\end{table}

\subsection{UNMT ablation study: Pre-training and bitext mining corpora}
\label{ssec:ablation}

In Sec.~\ref{ssec:unmt-results}, we mined bitext from the October 2019 Wikipedia snapshot whereas the pre-trained XLM embeddings were created prior to January 2019. Hence, it is possible that the UNMT BLEU increase would be smaller if the bitext were mined from the same corpus used for pre-training. We ran an ablation study to show the effect (or lack thereof) of the overlap between the pre-training and pseudo-parallel corpora.

For the En-Vi language pair, we used 5 million English and 5 million Vietnamese Wiki sentences to pre-train the XLM model. We only use text from the October 2019 Wiki snapshot. We mined 300k pseudo-parallel sentence pairs using our approach (Sec.~\ref{ssec:approach}) from the same Wiki snapshot. We created two datasets for XLM pre-training: a 10 million-sentence corpus that is disjoint from the 600k sentences of the mined bitext, and a 10 million-sentence corpus that contains all 600k sentences of the bitext. In Table \ref{table:ablation}, we show the BLEU increase on the IWSLT En-Vi task with and without using the mined bitext as parallel data, using each of the two XLM models as the initialization.

The benefit of using pseudo-parallel text is very clear; even if the pre-trained XLM model saw the pseudo-parallel sentences during pre-training, using mined bitext still significantly improves UNMT performance (23.1 vs.~28.3 BLEU). In addition, the baseline UNMT performance without the mined bitext is similar between the two XLM initializations (23.1 vs.~23.2 BLEU), which suggests that removing some of the parallel text present during pre-training does not have a major effect on UNMT.

\begin{table}[h!]
	\centering
	\footnotesize
\begin{tabu}{lcc}
\toprule
 & \textbf{w/o PP as bitext} & \textbf{w/ PP as bitext} \\
\midrule
XLM excl. PP text & 23.2 & 28.9 \\
XLM incl. PP text & 23.1 & 28.3 \\
\bottomrule
\end{tabu}
\caption{Tokenized UNMT BLEU scores on IWSLT'15 English-Vietnamese (tst2013) with XLM initialization. We mined 300k pseudo-parallel (PP) sentence pairs from En and Vi Wikipedia (Oct.\ 2019). We created two XLM models, with the pre-training corpus including or excluding the PP pairs. We compare their downstream UNMT performance with and without PP pairs as ``bitext'' during UNMT training.}
\label{table:ablation}
\end{table}

Finally, we trained a standard encoder-decoder model on the 300k pseudo-parallel pairs only, using the same Sockeye recipe in Sec.~\ref{ssec:unmt-results}. This yielded a BLEU score of 27.5 on En-Vi, which is lower than the best XLM-based result (i.e., 28.9), which suggests that the XLM initialization improves unsupervised NMT. A similar outcome was also reported in \citet{Lample2019XLM}.
\section{Related work}

\subsection{Parallel sentence mining}

Approaches to parallel sentence (or bitext) mining have been historically driven by the data requirements of statistical machine translation. Some of the earliest work in mining the web for large-scale parallel corpora can be found in \citet{resnik1998parallel} and \citet{resnik2003web}.  Recent interest in the field is reflected by new shared tasks on parallel extraction and filtering \citep{zweigenbaum-etal-2017-overview, koehn-etal-2018-findings} and the creation of massively-multilingual parallel corpora mined from the web, like WikiMatrix \citep{Schwenk2019WikiMatrix} and CCMatrix \citep{Schwenk2019CCMatrix}.

Existing parallel corpora have been exploited in many ways to create sentence representations for supervised bitext mining. One approach involves a joint encoder with a shared wordpiece vocabulary, trained as part of multiple encoder-decoder translation models on parallel corpora \citep{schwenk-2018-filtering}. \citet{Artetxe2019LASER} apply this approach at scale, and shared a single encoder and joint vocabulary across 93 languages. Another approach uses negative sampling to align the encoders' sentence representations for nearest-neighbor retrieval \citep{gregoire-langlais-2018-extracting, guo-etal-2018-effective}.

However, these approaches require training with initial parallel corpora. In contrast, \citet{Hangya2018Unsupervised} and \citet{hangya-fraser-2019-unsupervised} proposed unsupervised methods for parallel sentence extraction that use bilingual word embeddings induced in an unsupervised manner. Our work is the first to explore using contextual representations (mBERT; \citealp{devlin-etal-2019-bert}) in an unsupervised manner to mine for bitext, and to show improvements over the latest UNMT systems \citep{Lample2019XLM, Song2019MASS}, for which transformers and encoder/decoder pre-training have doubled or tripled BLEU scores on unsupervised WMT'16 En-De since \citet{Artetxe2018UNMT} and \citet{lample-etal-2018-phrase}.

\subsection{Self-training techniques}

Self-training refers to techniques that use the outputs of a model to provide labels for its own training. \citet{yarowsky1995unsupervised} proposed a semi-supervised strategy where a model is first trained on a small set of labeled data and then used to assign pseudo-labels to unlabeled data. Semi-supervised self-training has been used to improve sentence encoders that project sentences into a common semantic space. For example, \citet{clark2018semi} proposed cross-view training (CVT) with labeled and unlabeled data to achieve state-of-the-art results on a set of sequence tagging, MT, and dependency parsing tasks.

Semi-supervised methods require some annotated data, even if it is not directly related to the target task. Our work is the first to apply \emph{unsupervised} self-training for generating cross-lingual sentence embeddings.  The most similar approach to ours is the prevailing scheme for unsupervised NMT \cite{lample-etal-2018-phrase}, which relies on multiple iterations of backtranslation \cite{sennrich-etal-2016-improving} to create a sequence of pseudo-parallel sentence pairs with which to bootstrap an MT model.

\section{Conclusion}

In this work, we describe a novel approach for state-of-the-art unsupervised bitext mining using multilingual contextual representations. We extract pseudo-parallel sentences from unaligned corpora to create models that achieve state-of-the-art performance on unsupervised and low-resource translation tasks. Our approach is complementary to the improvements derived from initializing MT models with pre-trained encoders and decoders, and helps narrow the gap between unsupervised and supervised MT. We focused on mBERT-based embeddings in our experiments, but we expect unsupervised self-training to improve the unsupervised bitext mining and downstream UNMT performance of other forms of multilingual contextual embeddings as well.

Our findings are in line with recent work showing that multilingual embeddings are very useful for cross-lingual zero-shot and zero-resource tasks. Even without using aligned corpora, mBERT can embed sentences across different languages in a consistent fashion according to their semantic content. More work will be needed to understand how contextual embeddings discover these cross-lingual correspondences.

\iftaclpubformat

\section*{Acknowledgments}
We would like to thank the anonymous reviewers for their thoughtful comments.

\else
\fi

\bibliography{paper,anthology}
\bibliographystyle{acl_natbib}

\end{document}